\pdfoutput=1
\documentclass[conference]{IEEEtran}
\usepackage{blindtext, graphicx}
\usepackage{geometry}
\usepackage{subcaption}
\usepackage{subfig}
\usepackage{listings}
\usepackage{mathrsfs}
\usepackage{gensymb}
\usepackage{float}
\usepackage{cite}

\begin{document}

\title{Identifying Metastases in Sentinel Lymph Nodes with Deep Convolutional Neural Networks}

\author{\IEEEauthorblockN{Richard Chen}
\IEEEauthorblockA{Proscia Inc.}
\and
\IEEEauthorblockN{Yating Jing}
\IEEEauthorblockA{Proscia Inc.}
\and
\IEEEauthorblockN{Hunter Jackson}
\IEEEauthorblockA{Proscia Inc.}
}

\maketitle

\begin{abstract}
Metastatic presence in lymph nodes is one of the most important prognostic variables of breast cancer. The current diagnostic procedure for manually reviewing sentinel lymph nodes, however, is very time-consuming and subjective. Pathologists have to manually scan an entire digital whole-slide image (WSI) for regions of metastasis that are sometimes only detectable under high resolution or entirely hidden from the human visual cortex. From October 2015 to April 2016, the International Symposium on Biomedical Imaging (ISBI) held the Camelyon Grand Challenge 2016 to crowd-source ideas and algorithms for automatic detection of lymph node metastasis. Using a generalizable stain normalization technique and the Proscia Pathology Cloud computing platform, we trained a deep convolutional neural network on millions of tissue and tumor image tiles to perform slide-based evaluation on our testing set of whole-slide images images, with a sensitivity of 0.96, specificity of 0.89, and AUC score of 0.90. Our results indicate that our platform can automatically scan any WSI for metastatic regions without institutional calibration to respective stain profiles.

\end{abstract}

\IEEEpeerreviewmaketitle

\section*{Introduction}
\subsection*{Digital Pathology}
Pathology is a 150-year-old medical specialty \cite{11} that has seen a paradigm shift over the past few years with the advent of digital pathology. The digitization of tissue slides introduces a plethora of opportunities to leverage computer-assisted technologies to aid pathologists in diagnosing cancer\cite{12}. While proliferation of digital pathology is at an all time high, the industry has not crossed the rubicon into clinical diagnostics due to lack of standardization of image formats, system noise, and lack of clinical and technical studies on digital pathology systems.

The inherent problem in pathology is subjectivity. The discipline is plagued with human variability from tissue acquisition, improper staining techniques, and subjectivity in diagnosing under a microscope. A pathologist looks for patterns in a tissue sample and uses his/her medical training to interpret those patterns and make a diagnosis \cite{1}. As evidenced through many applications in myriad industries, computer-assisted pattern recognition software can match or even supersede a human's ability to recognize patterns \cite{8}.

Though digital pathology is on the precipice of wide-spread adoption, the difficulties in hardware scanning variability and fear of "black-box" computational tools has lead to a longer adoption curve than seen in other medical specialties that have gone totally digital, e.g. radiology \cite{9}.

Region of Interest (ROI) detection algorithms, like the one proposed here have the potential to act as an intermediary clinical decision support tool, rather than immediately moving to a fully computerized diagnostic procedure \cite{7}. ROI detection tools cut down time for analyzing whole slide tissue sections by reducing the signal-to-noise ratio and directing the pathologist to those regions that are flagged for containing certain properties indicative of the region of interest, as learned by the model. 

While computerized primary diagnosis is still in its nascent stages, these such tools provide an expedient route to enabling precision medicine in pathology by automating the ROI detection and alleviating diagnostic subjectivity.

\subsection*{Camelyon Competition}
From October 2015 to April 2016, the International Symposium on Biomedical Imaging (ISBI) held the Camelyon Grand Challenge 2016 to crowd-source ideas and algorithms for automatic detection of lymph node metastasis\cite{camelyon16}\cite{Beck}. The following two metrics in the challenge were used to evaluate the performance of the algorithms:

\begin{enumerate}
\item Slide-based Evaluation: algorithm performance on discriminating between normal slides and metastasis slides
\item Lesion-based Evaluation: algorithm performance on lesion detection and localization
\end{enumerate}

The dataset in the challenge contains a total of 400 whole-slide images (WSIs) of sentinel lymph node from two independent datasets collected in Radboud University Medical Center (UMC) (Nijmegen, the Netherlands) and the UMC Utrecht (Utrecht, the Netherlands). Whole-slide images are large image files organized in a multi-resolution pyramid structure, in which each image in the pyramid is a downsampled version of the highest resolution image. The first training dataset consists of 170 WSIs of lymph node (100 Normal and 70 Tumor) and the second training dataset consists of 100 WSIs (60 Normal and 40 Tumor). The first testing dataset consists of 80 WSIs and the second testing dataset consists of 50 WSIs. The ground truth data of the training set came from a pathologist who manually drew contours of regions of lymph node metastasis.

Because the labeled data for the test set was never made public in the Camelyon Challenge, we performed a 60-20-20 train-validation-test split on the 270 WSIs provided as training data. In addition, we only tested the performance of our algorithms using slide-based evaluation. Our results are based on a small subset of the entire image set; however, these initial results are motivation for further work on the entire dataset and completion of the Camelyon challenge.

\section*{Method}
Our framework for detecting metastases in sentinel lymph nodes can be modularized into four components: image preprocessing and tiling of WSIs, color deconvolution and stain normalization of WSI tiles, tile-based classification using convolutional neural networks (CNNs), and post-processing of tumor probability heatmaps of the WSIs. The libraries and scientific packages used were OpenSlide, NumPy, Pillow, OpenCV, Caffe, and pyCaffe.

\subsection*{Image Preprocessing \& Tiling}
\begin{figure}[htp]
    \begin{subfigure}{0.15\textwidth}
      \centering
      \includegraphics[width=\linewidth]{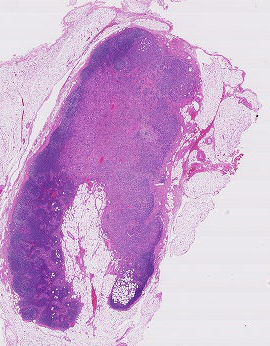}
      \caption{}
      \label{fig:sfig1}
    \end{subfigure}
    \begin{subfigure}{0.15\textwidth}
      \centering
      \includegraphics[width=\textwidth]{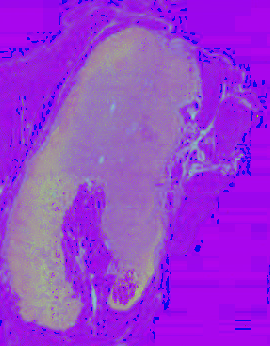}
      \caption{}
      \label{fig:sfig2}
    \end{subfigure}
    \begin{subfigure}{0.15\textwidth}
      \centering
      \includegraphics[width=\textwidth]{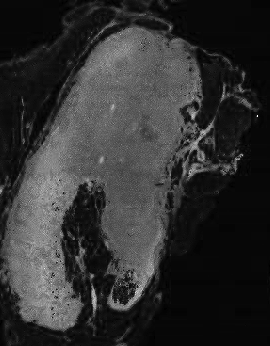}
      \caption{}
      \label{fig:sfig3}
    \end{subfigure}\\[-1ex]
    \begin{subfigure}{0.15\textwidth}
      \centering
      \includegraphics[width=\textwidth]{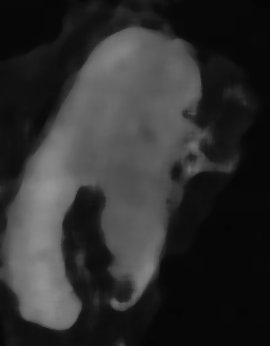}
      \caption{}
      \label{fig:sfig4}
    \end{subfigure}
    \begin{subfigure}{0.15\textwidth}
      \centering
      \includegraphics[width=\textwidth]{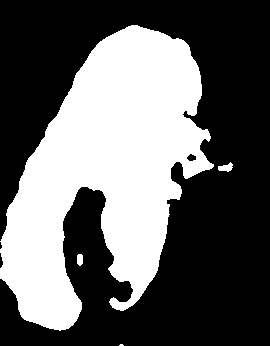}
      \caption{}
      \label{fig:sfig5}
    \end{subfigure}
    \begin{subfigure}{0.15\textwidth}
      \centering
      \includegraphics[width=\textwidth]{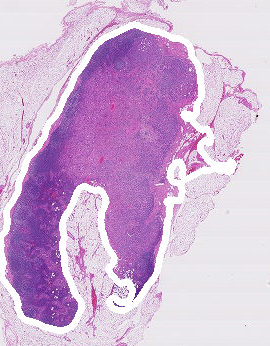}
      \caption{}
      \label{fig:sfig6}
    \end{subfigure}\\[-2ex] 
    \caption{Image preprocessing steps for tissue segmentation. (a) Original Image (b) HSV colorspace (c) Saturation channel (d) Median blur (e) Otsu's Binarization (e) Original Image with contour overlay}
\end{figure}

\begin{figure*}
  \includegraphics[width=\textwidth]{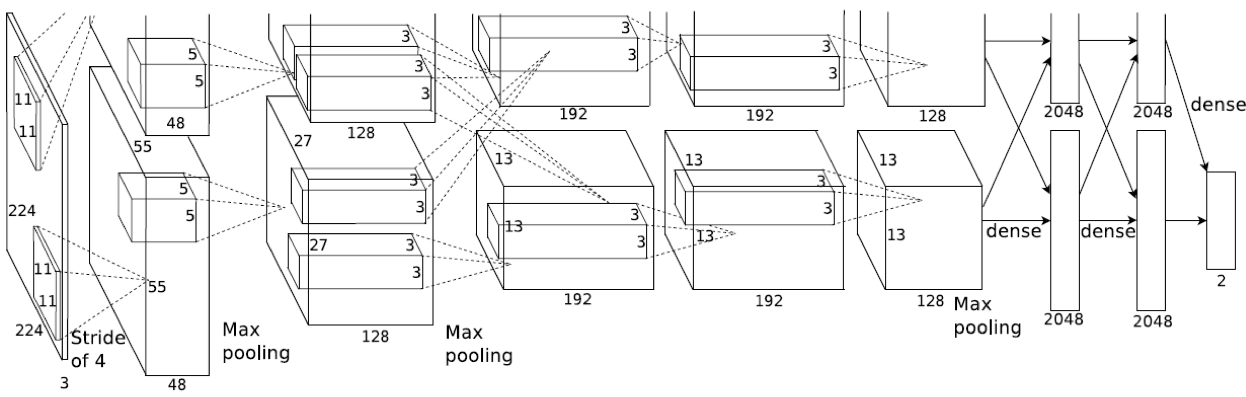}
  \caption{AlexNet Layer Architecture}
\end{figure*}

For each WSI, we perform a connected components analysis to draw contours around only tissue regions of the slide, cutting down total pixels processed by over 70\%. Specifically, we transformed the WSI's RGB color vector to the HSV colorspace and performed Otsu's Binarization on only the saturation channel to separate brightly-stained tissue regions from the gray space of the background region. We also used a median blur to filter out fatty tissue and background artifacts. To reduce computational run-time, we performed all of our segmentation on a downsampled version of the original image and scaled the pixel coordinates of the contour regions to the dimension with the highest resolution. Specifically, we performed segmentation on the thumbnail dimension of all the WSI's. Using OpenSlide\cite{openslide}, we were able to compute level-based dimensions which are used to create a linear scaling function to scale contours created at any resolution to be represented in a higher dimensional resolution space without losing any contour information. Knowing the downsample factor from the thumbnail dimension to the level 1 dimension of the WSI, we perform a scale transformation and rapidly calculate the contours of the tissue region at the highest spatial resolution of the image. Finally, within each contour region, we partitioned the tissue and tumor regions into 256 x 256 tiles at 40x (level 1) using a fully parallel computing architecture in the Pathology Cloud Computing Platform that expedites the process of creating the training set. 

\subsection*{Stain Normalization}

Since the Camelyon challenge dataset was generated from two separate institutions, we wanted to eliminate any stain variability that could negatively affect training and testing. By performing stain normalization, we can perform a non-linear color feature mapping that maps all of the respective training images by to a target stain. Without elimination of the stain variability, the heterogeneity of the stains will introduce bias to our model, inducing a bimodal stain vector color distribution. We performed stain normalization on each 256 x 256 tile generated during WSI tiling.

\begin{figure}[H]
    \begin{subfigure}{0.15\textwidth}
      \centering
      \includegraphics[width=\linewidth]{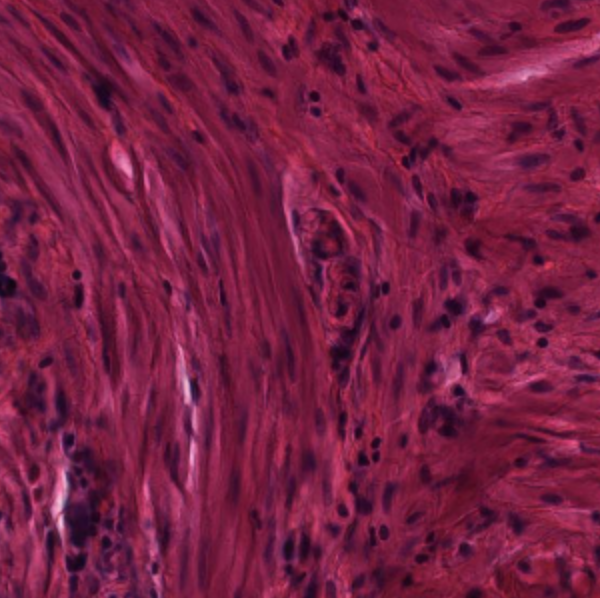}
      \caption{}
      \label{fig:sfig1}
    \end{subfigure}
    \begin{subfigure}{0.15\textwidth}
      \centering
      \includegraphics[width=\textwidth]{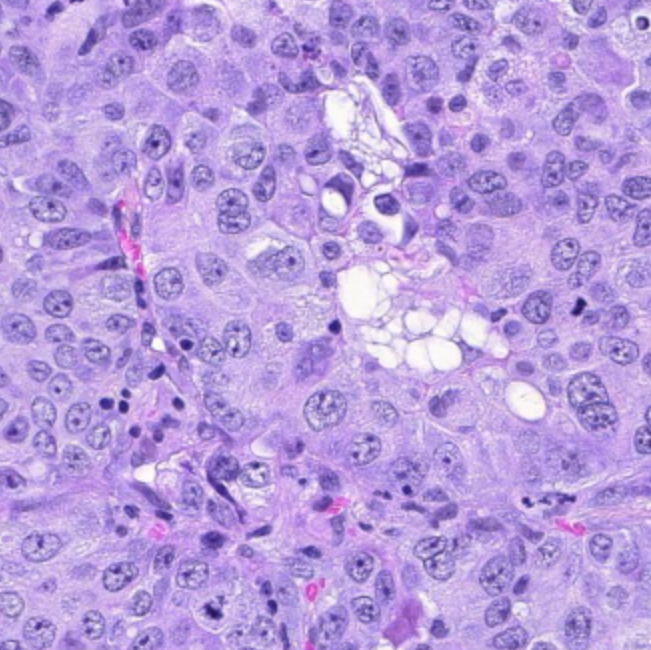}
      \caption{}
      \label{fig:sfig2}
    \end{subfigure}
    \begin{subfigure}{0.15\textwidth}
      \centering
      \includegraphics[width=\textwidth]{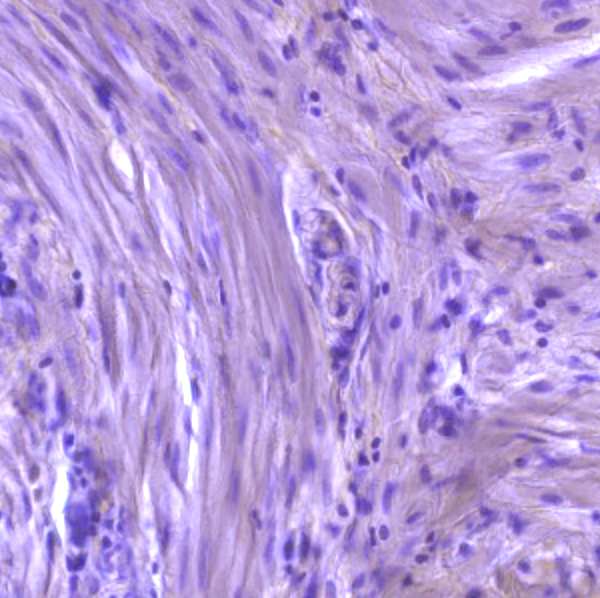}
      \caption{}
      \label{fig:sfig3}
    \end{subfigure}\\[-2ex]
    \caption{An example of stain normalization on a heavily overstained source image (a) Source Image (b) Target Image (c) Output Image}
\end{figure}

\subsection*{Tile-Based Classification using Convolutional Neural Networks}
After creating a training set of 256 x 256 tiles, we trained a convolutional neural network to discriminate between tissue and tumor tiles. Namely, we chose AlexNet\cite{3}. The architecture of AlexNet consists of five convolutional and three fully-connected learned layers, 60 million parameters, and a two-way softmax to assign probabilities to the tissue and tumor class labels. In addition to stain normalization, we preprocessed each training image by subtracting the mean activity over the training set from each pixel. We also extracted random 224 x 224 patches and their horizontal reflections from each 256 x 256 tile in our training set and trained our model on these patches to prevent overfitting.

\subsection*{Post-Processing of Tumor Probability Heatmap}
Given a WSI in the Camelyon training data, we generated its corresponding tumor probability heatmap, with each pixel assigned a value $p$ for $p \in [0,1]$, indicating the probability of metastasis. To analyze these heatmaps efficiently, we generated tumor probability heatmaps at level 4 instead of level 0 due to computational complexity. To accomplish this, for every tile within the tissue and contour regions at level 0, we used our AlexNet model to assign a probability to each tile, and color-mapped that tile's probability to its corresponding downsampled tile at level 4. In the pyramidal structure of each WSI, the image at level 4 is 8 downsamples greater than the image at level 0. Specifically, for a 256 x 256 tile at level 4, it would take 64 256 x 256 tiles at level 0 to represent the same region in the image. As a result, probabilities assigned to 256 x 256 tiles at level 1 can be mapped and colored to 32 x 32 pixel regions at level 4. After generating the tumor probability heatmap, we performed post-processing for slide-based evaluation.

\begin{figure}[htp]
    \begin{subfigure}{0.225\textwidth}
      \centering
      \includegraphics[width=\linewidth]{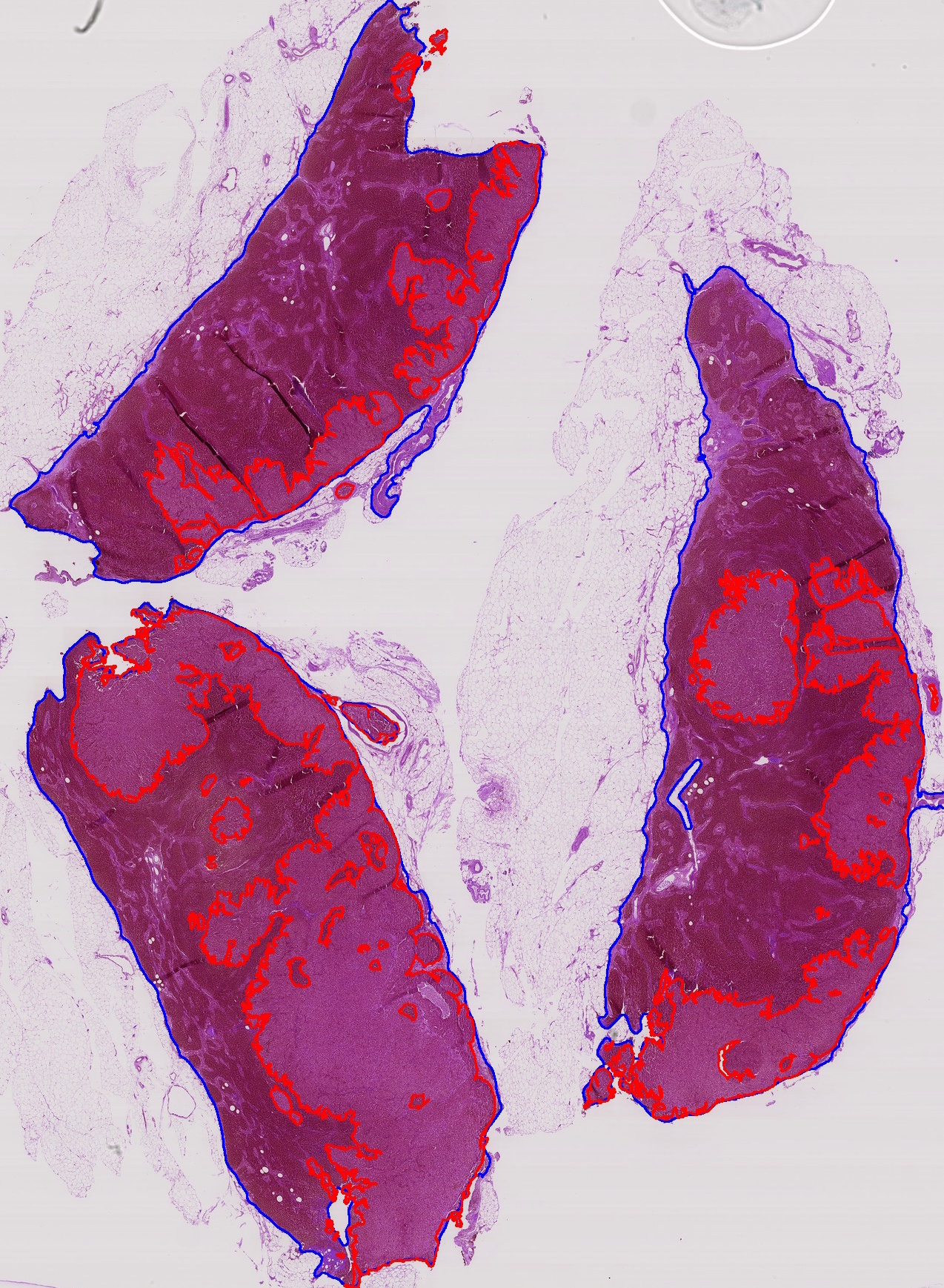}
      \caption{}
      \label{fig:sfig1}
    \end{subfigure}
    \begin{subfigure}{0.225\textwidth}
      \centering
      \includegraphics[width=\textwidth]{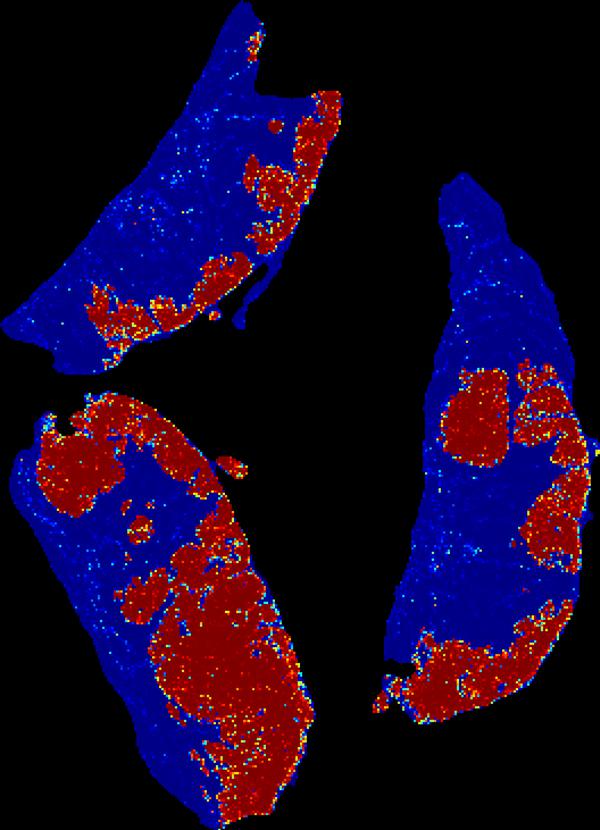}
      \caption{}
      \label{fig:sfig2}
    \end{subfigure}\\[-2ex]
    \caption{A comparison between the (a) ground truth data of Tumor 087 and (b) tumor probability heatmap of Tumor 087}
\end{figure}

For slide-based evaluation, a 60-20-20 train-validation-test split was done on 270 tumor probability heatmaps generated from the WSIs provided as training data, with each partition having an approximately equal proportion of tumor to tissue tiles. For each heatmap, we computed geometrical and morphological features about its tumor probability distribution. Such features include:  max, mean, variance, skewness, and kurtosis of (area, perimeter, compactness, rectangularity, solidity of tumor regions), average prediction across tumor region, total number of tumor regions, and the total number pixels with probability greater than 0.90. After computing these features, we trained a Random Forest classifier, and used it to evaluate our test set of WSIs. 

\section*{Experimental Results}
We first evaluated tile-based classification accuracy with and without stain normalization. Under this evaluation metric, where we were able to accurately classify all given tiles in the test set as belonging to a tissue region or a metastatic region with 92.7\% without stain normalization, and 96.6\% accuracy with stain normalization. Because we achieved a higher tile-based classification accuracy under stain normalization, we performed slide-based evaluation with this extra pre-processing step.

For slide-based evaluation, the Random Forest classifier yielded an average image-based classification sensitivity of .96, average specificity of .89, and a computed area under the Receiver Operating Characteristic Curve (AUC) score of 0.90. 

\section*{Conclusion}
The results of the study show that stain normalization is a crucial part of building a generalizable deep learning model for identifying metastatic regions in sentinel lymph nodes, as it eliminates the variability induced by different stains from the two locations. This process significantly improved classification performance of our model and indicates that this model could be extended to any breast cancer digitized images without institutional recalibration. Our algorithms and model will be subject to additional training and validation on the full image set provided by ISBI and the Camelyon Challenge 2016. We are grateful to the organizers and all those involved in the competition, and we look forward to improving our results and computing a tumor localization score for the fully trained model.

\bibliographystyle{unsrt}
\bibliography{bibliography}

\end{document}